\documentclass[10pt,twocolumn,letterpaper]{article}

\usepackage[pagenumbers]{wacv} 

\usepackage{algorithm}
\usepackage{algpseudocode}
\usepackage{multirow}

\usepackage{multibib}
\newcites{supp}{Supplementary References}

\definecolor{wacvblue}{rgb}{0.21,0.49,0.74}
\usepackage[pagebackref,breaklinks,colorlinks,allcolors=wacvblue]{hyperref}

\def\wacvPaperID{1975} 
\def\confName{WACV}
\def\confYear{2027}

\title{MIME: Multimodal Interactive Motion Encoder}

\author{Addison Zucek, Prerit Gupta, Kamila Kuatova, Aniket Bera\\
Purdue University\\
{\tt\small \{azucek, gupta596, kkuatova, aniketbera\}@purdue.edu}
}

\begin{document}
\maketitle
\begin{abstract}
Text-motion representation learning has advanced rapidly, with growing interest in multi person interactions for animation, AR/VR, and embodied AI. These settings require representations that align language with both individual actor dynamics and the relationships between actors. We introduce the Multimodal Interactive Motion Encoder (MIME), which, to our knowledge, represents the first dedicated multimodal encoder designed specifically for two person interactive motion. MIME captures individual and shared structure using stream based co-attention with explicit interaction features and curriculum based contrastive training. On Inter-X text-motion retrieval, MIME consistently outperforms early and late fusion baselines across gallery sizes, achieving a 12.8\% relative improvement in text-to-motion R@1 at a 2,000-sample gallery. We further evaluate MIME as a frozen auxiliary prior within TIMotion and InterMask on the unseen InterHuman dataset. MIME improves semantic alignment metrics while maintaining comparable FID in TIMotion. These results show that interaction aware multimodal encoding improves multi person motion retrieval and transfers across datasets to support downstream motion generation.
\end{abstract}
    
\section{Introduction}
\label{sec:intro}

\begin{figure*}[tb]
\centering
\includegraphics[width=\linewidth]{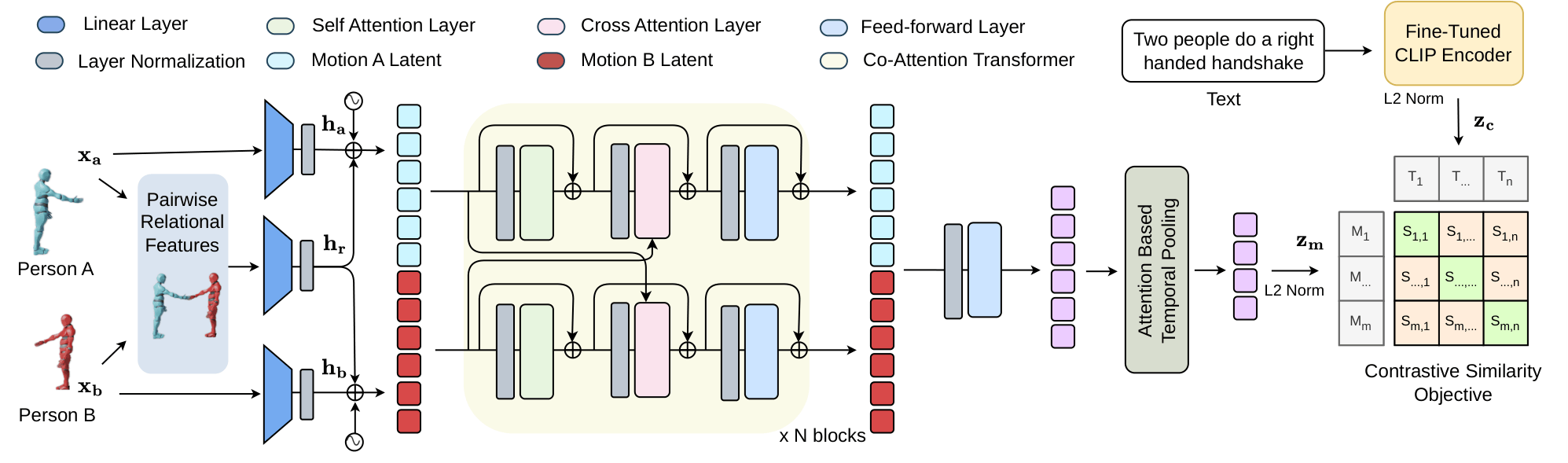}
\caption{MIME architecture. Motion pairwise features are calculated and injected into both streams $h_a, h_b$. The streams are passed through N blocks comprised of self+cross attention and a feed forward layer. Then the embedding is compressed to an appropriate size for contrastive learning with the text query embedding.}
\label{fig:MIME_Arc}
\end{figure*}

Human motion generation has made substantial progress, driving new capabilities in animation~\cite{azadi2023make, yi2024generating, peng2018deepmimic}, AR/VR~\cite{patel2025uniegomotion, zhang2022egobody}, and robotics~\cite{rempe2026kimodo, zhang2026learning}. Research on generating motion for a single person has matured significantly, but many real-world scenarios involve people interacting with one another. Consequently, text-guided interactive motion synthesis has emerged as a critical research direction~\cite{li2024twoinoneunifiedmultipersoninteractive, fan20253d}. A requirement for these systems is a discriminative shared text-motion latent space where interactive motion sequences and their semantic captions are closely aligned. High-quality representations in this space are essential for cross-modal retrieval~\cite{wang2025cross}, dataset construction and text-guided generation, but existing multimodal encoders remain less capable for interactive scenarios.

Complex interactions such as hugging or dancing require modeling intricate spatial and temporal relationships, including body synchronization and contact timing. Standard early-fusion concatenation feeds multiple actors into a single temporal model, forcing the network to implicitly derive relative geometry from flattened representations. Conversely, late-fusion approaches defer interaction modeling until after temporal processing, missing frame-synchronous relational cues. Both paradigms can struggle to preserve actor-specific structure, limiting alignment in the shared latent space.

To address this gap, we introduce the Multimodal Interactive Motion Encoder (MIME). Instead of directly concatenating multiple motions into a single transformer, MIME employs a co-attention architecture across separate actor streams. This allows the network to preserve individual motion structure while explicitly modeling how the subjects interact. We further enrich these representations with explicit frame-level relational features and utilize a lightweight curriculum schedule~\cite{bengio2009curriculum} during contrastive training. Optimized via a symmetric contrastive retrieval objective, MIME maps paired multi-person motions and text descriptions to the same latent region. We evaluate our approach on the Inter-X dataset~\cite{xu2024inter}.

To test the generalizability of the learned latent space, we conduct downstream validation on the InterHuman dataset. Deployed as a frozen semantic-conditioning probe within two generation pipelines, MIME improves retrieval-based text-motion alignment in both, and in TIMotion~\cite{wang2025TIMotion} it maintains comparable FID.

In summary, the main contributions are as follows:

\begin{enumerate}
\item We introduce MIME, a dedicated text-motion representation learning framework for two-person interactive motion that, to the best of our knowledge, is the first to jointly encode text and dyadic motion interactions.

\item We develop an interaction-aware encoding architecture that captures both individual motion semantics and inter-person dependencies through bidirectional co-attention, explicit relational features and curriculum-guided contrastive learning.

\item Through extensive retrieval experiments, ablation studies \& downstream task evaluations, we show that MIME learns a more discriminative interaction representation and can serve as an effective frozen prior for improving text-motion alignment in interactive motion generation.

\end{enumerate}

\section{Related Work}
\label{sec:related}

\noindent\textbf{Multimodal Text-Motion Encoding.} 
Learning a shared latent space~\cite{ngiam2011multimodal, plappert2016kit, punnakkal2021babel, tevet2022motionclip} between natural language and human motion is fundamental to cross-modal retrieval, motion editing dataset construction, and generative conditioning. TMR~\cite{petrovich2023tmr} directly formulates text-to-motion retrieval as contrastive alignment between motion and language embeddings while other more recent approaches~\cite{li2025lamp, ren2025wamo, zhang2026sgar} further refine motion-language encoders for fine-grained semantic correspondence. However, these methods are primarily developed for single-person motion and do not explicitly target reusable two-person interaction encoders. Straightforward multi-person adaptations typically rely on early or late fusion, whereas MIME preserves actor-specific streams and models frame-synchronous dependencies through explicit interaction features and bidirectional co-attention.

\noindent\textbf{Text-Driven Motion Generation and Editing.}
Text-conditioned human motion generation has expanded rapidly through diffusion models, transformers, and masked modeling. Early approaches explored both continuous latent-variable formulations and diffusion-based generation~\cite{petrovich2022temos, guo2022generating, tevet2022human}. Subsequent methods have improved generation fidelity, controllability, and semantic alignment through discrete motion representations, autoregressive modeling, retrieval augmentation, masked prediction, and explicit spatial control~\cite{guo2024momask, liang2024intergen, xie2024omnicontrol, meng2025rethinking, zhang2023generating, jiang2023motiongpt, zhang2023remodiffuse, van2017neural}. In parallel, motion editing takes source motions and natural language instructions as input to produce modified sequences. Motion editing frameworks~\cite{athanasiou2024motionfix, li2025simmotionedit} highlight the importance of robust text-motion representations for retrieval, semantic evaluation, and editing supervision. These works motivate expressive text-motion representations, but they do not directly answer the representation question studied here: how to build a reusable multimodal encoder for paired human-human interactions.

\noindent\textbf{Interactive Human Motion Modeling.}
Interactive motion introduces relational constraints absent from isolated motion, including relative position, synchronization, contact timing, role structure, and collision-free coordination~\cite{ghosh2024remos, li2024duolando, sui2026survey, xu2025multi}. Large-scale interaction datasets~\cite{xu2024inter, liang2024intergen, Gupta_2025_ICCV} have catalyzed research into two-person dynamics. Recent generation frameworks~\cite{wang2025TIMotion, gupta2025unified} show that temporal modeling and interaction mixing should be handled directly rather than treated as a simple extension of single-person motion. While models like TIMotion embed interaction modeling within end-to-end generative pipelines, MIME focuses on a complementary problem: learning a reusable, retrieval-focused multimodal encoder for interactive motion. Motivated by the broader principle that cross-stream attention can model dependencies between structured inputs~\cite{lu2016hierarchical, lu2019vilbert, tsai2019multimodal}, MIME preserves actor-specific streams while introducing explicit interaction features and bidirectional co-attention. We therefore position MIME not as a competing generation framework, but as an interaction-aware representation that can support retrieval, semantic evaluation, and downstream conditioning.
\section{Methods}
\label{sec:methods}

\subsection{Problem Formulation}
\label{sec:problem_formulation}

Given person A's motion $\mathbf{x_a}=\{x_a^i\}_{i=1}^{N}$, person B's motion $\mathbf{x_b}=\{x_b^i\}_{i=1}^{N}$, and a natural-language description $c$ of their dyadic interaction, our objective is to learn a shared text--motion embedding space in which corresponding descriptions and motion pairs are close to one another. Here, $N$ denotes the number of frames, and $x_a^i$ and $x_b^i$ represent the motions of person A and B at frame $i$.

A motion encoder maps the paired sequence to an interaction embedding, $\mathbf{z_m}=f_{\mathrm{m}}(\mathbf{x_a},\mathbf{x_b})$ while a text encoder maps the description to $\mathbf{z_c}=f_{\mathrm{text}}(c)$. The encoders are trained such that matched text--motion pairs have higher similarity than mismatched pairs. This shared space supports both retrieval directions. Given a text query $c_i$, text-to-motion retrieval ranks a gallery of motion pairs
$\{(\mathbf{x_a}_j,\mathbf{x_b}_j)\}_{j=1}^{G}$ to retrieve
$(\mathbf{x_a}_i,\mathbf{x_b}_i)$. Conversely, motion-to-text retrieval ranks candidate descriptions for a query motion pair and retrieves its corresponding caption. The learned text representation can also condition a generative model to synthesize semantically aligned and mutually coordinated two-person motion.

\paragraph{Human Motion Representation.}
Each frame of a person's motion is represented as
$x^i=[\Delta p,\theta_{\mathrm{body}},\theta_{\mathrm{root}}]$, where
$x^i\in\mathbb{R}^{135}$. Here, $\Delta p\in\mathbb{R}^{3}$ denotes the root translation displacement between successive frames, while $\theta_{\mathrm{body}}\in\mathbb{R}^{6(N_j-1)}$ and $\theta_{\mathrm{root}}\in\mathbb{R}^{6}$ denote the 6D rotation representations of the body joints and the root joint, respectively. We use the SMPL model \cite{SMPL:2015} with $N_j = 22$ joints. The 6D rotation representation avoids the discontinuities associated with Euler angles and provides a stable representation for learning temporal human motion.

\subsection{Model Architecture}
\label{sec:model_architecture}

Our framework consists of a text encoder and an interactive motion encoder that project the interaction description and the paired human motions into a shared embedding space.

\subsubsection{Text Encoder}
Given an interaction description $c$, the text encoder maps it to a latent representation $\mathbf{z_c} = f_{\mathrm{text}}(c)\in \mathbb{R}^{D}$
where $D$ denotes the shared embedding dimension. We use the CLIP text encoder and project into the shared latent space~\cite{radford2021learning}. For all models and experiments the text encoder is fine-tuned during
training, allowing the language representation to adapt toward interaction-specific motion semantics. The final text embedding is L2-normalized.

\subsubsection{Interactive Motion Encoder}
The interactive motion encoder $\mathbf{z_m}=f_{\mathrm{m}}(\mathbf{x_a},\mathbf{x_b})$ jointly captures the temporal dynamics of each person and the coordination between them through role-aware motion projection, relational feature injection, person-specific self-attention, bidirectional co-attention transformers and attention-based temporal pooling.

\paragraph{Role-Aware Motion Projection.}
We first project the frame-level motion features of each person into a shared
$D$-dimensional latent space:
\begin{equation}
    h_a^i
    =
    \mathrm{LN}_a
    \left(
        W_a x_a^i + b_a
    \right)
    +
    e_a
    +
    \gamma^i,
\end{equation}
\begin{equation}
    h_b^i
    =
    \mathrm{LN}_b
    \left(
        W_b x_b^i + b_b
    \right)
    +
    e_b
    +
    \gamma^i,
\end{equation}
where $W_a$ and $W_b$ are learned projection matrices, $b_a$ and $b_b$ are
bias terms, $\mathrm{LN}_a$ and $\mathrm{LN}_b$ denote person-specific layer
normalization operations, and $\gamma^i$ is the sinusoidal positional encoding of
frame $i$. The learnable person-type embeddings $e_a$ and $e_b$ are randomly
initialized and explicitly distinguish the two motion streams, allowing the
encoder to preserve their asymmetric interaction roles.

\paragraph{Relational Feature Encoding.}
To provide MIME with explicit information about the spatial relationship between the two individuals, we augment each person's base motion representation with four root-relative features as follows: 

\begin{equation} \tilde{x}_{a}^i = \left[ x_{a}^i;\, \left\|p_a^i-p_b^i\right\|_2;\, p_a^i-p_b^i \right] \in \mathbb{R}^{139}. \label{eq:relational-features4} \end{equation} \begin{equation} \tilde{x}_{b}^i = \left[ x_{b}^i;\, \left\|p_b^i-p_a^i\right\|_2;\, p_b^i-p_a^i \right] \in \mathbb{R}^{139}. \label{eq:relational-features4.2} \end{equation}
Here, $p_{{a,b}}^i \in \mathbb{R}^{3}$ denotes the 3D root position of person $a$ or $b$ at frame $i$. The augmented features consist of distance between the root positions of both persons and the signed root displacement along the three spatial axes.

To explicitly capture the frame-level relationship between the two people, we construct the relational feature
\begin{equation}
    r^i
    =
    \left[
        \tilde{x}_a^i;
        \tilde{x}_b^i;
        \tilde{x}_b^i-\tilde{x}_a^i;
        \rho^i
    \right],
    \label{eq:relational-feature}
\end{equation}
where the difference term $\tilde{x}_b^i-\tilde{x}_a^i$ encodes their relative configuration in the motion feature space. The scalar $\rho^i$ measures the relative motion magnitude of person A with respect to both people at frame $i$. Using their root translation displacements, we define
\begin{equation}
    \rho^i
    =
    \frac{
        \left\lVert \Delta p_a^i \right\rVert_2^2
    }{
        \left\lVert \Delta p_a^i \right\rVert_2^2
        +
        \left\lVert \Delta p_b^i \right\rVert_2^2
        +
        \epsilon
    },
\end{equation}
where $\epsilon$ is a small constant introduced for numerical stability. A value of $\rho^i$ close to $1$ indicates that person A has a larger root translation magnitude, while a value close to $0$ indicates that person B is moving more strongly.

The relational feature is then projected into the shared latent space:
\begin{equation}
    h_r^i
    =
    \mathrm{LN}_r
    \left(
        W_r r^i+b_r
    \right)
    +
    e_r,
\end{equation}
where $W_r$ and $b_r$ are learned projection parameters, $\mathrm{LN}_r$ denotes layer normalization, and $e_r$ is a learnable relation-type embedding. We incorporate this relational representation into both person-specific streams:
\begin{equation}
    \tilde{h}_a^i=h_a^i+h_r^i,
    \qquad
    \tilde{h}_b^i=h_b^i+h_r^i.
\end{equation}
This early relational fusion provides both motion streams with explicit information about the joint interaction before self-attention and cross-attention are applied.

\paragraph{Bidirectional Co-Attention Transformer.}

The two motion streams are processed by a stack of $L$ co-attention transformer layers. Each layer consists of person-specific self-attention, bidirectional cross-attention, and feed-forward networks, with pre-normalization and residual connections applied throughout.

Let $\mathbf{h_a^{\ell}}
    =
    \{h_{a}^{i,\ell}\}_{i=1}^{N}$ and
    $\mathbf{h_b^{\ell}}
    =
    \{h_{b}^{i,\ell}\}_{i=1}^{N}$
denote the latent motion sequences of persons A and B entering the $\ell$-th layer. For the first layer, we set
\begin{equation}
    \mathbf{h_a^{0}}=\{\tilde{h}_a^i\}_{i=1}^{N},
    \qquad
    \mathbf{h_b^{0}}=\{\tilde{h}_b^i\}_{i=1}^{N}.
\end{equation}

We first apply person-specific self-attention (SA) layer to independently model the temporal dynamics of each motion stream:
\begin{equation}
    \mathbf{\bar{h}_a^{\ell}}
    =
    \mathbf{h_a^{\ell}}
    +
    \mathrm{SA}_a^{\ell}
    \left(
        \mathrm{LN}_{a}^{\ell}
        \left(
            \mathbf{h_a^{\ell}}
        \right)
    \right),
\end{equation}
\begin{equation}
    \mathbf{\bar{h}_b^{\ell}}
    =
    \mathbf{h_b^{\ell}}
    +
    \mathrm{SA}_b^{\ell}
    \left(
        \mathrm{LN}_{b}^{\ell}
        \left(
            \mathbf{h_b^{\ell}}
        \right)
    \right).
\end{equation}

We then apply a bidirectional cross-attention (CA) layer to exchange information between the two persons. Person A attends to person B as
\begin{equation}
    \mathbf{\hat{h}_a^{\ell}}
    =
    \mathbf{\bar{h}_a^{\ell}}
    +
    \mathrm{CA}_{a \leftarrow b}^{\ell}
    \left(
        \mathrm{LN}_{a}^{\ell}
        \left(
            \mathbf{\bar{h}_a^{\ell}}
        \right),
        \mathrm{LN}_{b}^{\ell}
        \left(
            \mathbf{\bar{h}_b^{\ell}}
        \right),
        \mathrm{LN}_{b}^{\ell}
        \left(
            \mathbf{\bar{h}_b^{\ell}}
        \right)
    \right)
\end{equation}
where the three arguments denote the query, key, and value sequences, respectively. Similarly, person B attends to person A:
\begin{equation}
    \mathbf{\hat{h}_b^{\ell}}
    =
    \mathbf{\bar{h}_b^{\ell}}
    +
    \mathrm{CA}_{b \leftarrow a}^{\ell}
    \left(
        \mathrm{LN}_{b}^{\ell}
        \left(
            \mathbf{\bar{h}_b^{\ell}}
        \right),
        \mathrm{LN}_{a}^{\ell}
        \left(
            \mathbf{\bar{h}_a^{\ell}}
        \right),
        \mathrm{LN}_{a}^{\ell}
        \left(
            \mathbf{\bar{h}_a^{\ell}}
        \right)
    \right)
\end{equation}
Because cross-attention is computed over the complete temporal sequences, each frame of one person can attend to any valid frame of the other, enabling the model to capture both synchronized interactions and temporally delayed responses.

Finally, person-specific feed-forward networks (FFN) with GeLU Activation function \cite{hendrycks2016gaussian} update the two streams:
\begin{equation}
    \mathbf{h_a^{\ell+1}}
    =
    \mathbf{\hat{h}_a^{\ell}}
    +
    \mathrm{FFN}_a^{\ell}
    \left(
        \mathrm{LN}_{a}^{\ell}
        \left(
            \mathbf{\hat{h}_a^{\ell}}
        \right)
    \right),
\end{equation}
\begin{equation}
    \mathbf{h_b^{\ell+1}}
    =
    \mathbf{\hat{h}_b^{\ell}}
    +
    \mathrm{FFN}_b^{\ell}
    \left(
        \mathrm{LN}_{b}^{\ell}
        \left(
            \mathbf{\hat{h}_b^{\ell}}
        \right)
    \right).
\end{equation}

\paragraph{Frame-Wise Fusion.}

After the final co-attention layer, the representations of both people are concatenated at each frame:
\begin{equation}
    u^i
    =
    f_{\mathrm{fuse}}
    \left(
        \left[
            h_a^i;
            h_b^i
        \right]
    \right),
\end{equation}
where $f_{\mathrm{fuse}}$ is a frame-wise multilayer perceptron that projects the concatenated representation back to $D$ dimensions.

\paragraph{Attention-based Temporal Pooling.}

We then aggregate the frame-level interaction features using learnable query-based temporal pooling. Given a trainable query vector $q\in\mathbb{R}^{D}$, the importance of frame $i$ is computed as
\begin{equation}
    s^i
    =
    (u^i)^\top q.
\end{equation}
After masking padded frames, the normalized attention weight is
\begin{equation}
    \alpha^i
    =
    \frac{
        \exp(s^i)
    }{
        \sum_{j\in\mathcal{V}}
        \exp(s^j)
    },
\end{equation}
where $\mathcal{V}$ denotes the set of frames that are valid for both people. The final interaction-level motion representation is
\begin{equation}
    \mathbf{z_m}
    =
    \sum_{i\in\mathcal{V}}
    \alpha^i u^i.
\end{equation}
Finally, $\mathbf{z_m}$ is $\ell_2$-normalized and aligned with the text representation $\mathbf{z_c}$ through a contrastive learning objective.

\subsection{Training Objective}

Following TMR \cite{petrovich2023tmr}, given a minibatch of $n$ paired, L2-normalized motion and text embeddings
$\{(z_m^i,z_c^i)\}_{i=1}^{n}$, we compute similarities
$s_{ij}=\alpha (z_m^i)^\top z_c^j$, where $\alpha$ is a learned logit scale. We optimize the average of the motion-to-text and text-to-motion
cross-entropy losses by treating $(z_m^i,z_c^i)$ as the positive pair and
all other within-batch pairs as negatives.

\subsection{Curriculum Sampling}

To strengthen contrastive training, MIME uses a staged semantic batch sampler~\cite{wu2017sampling}. We first
compute anchor-text embeddings $a_i$ for the training set and form a text-space similarity
matrix
\begin{equation}
S_{ij}=a_i^\top a_j,
\end{equation}
with self-similarities masked. During warmup, batches are sampled uniformly. After warmup,
each batch is formed by selecting a random anchor and sampling the remaining examples from a
window over its neighbors sorted by $S_{ij}$. Once an example is placed in a batch, it is
removed from the candidate pool for the rest of that epoch, which encourages broad
training-set coverage while increasing the semantic difficulty of within-batch negatives.

The window location is controlled by a curriculum hardness parameter
\begin{equation}
\eta(e)=\eta_{\max}\frac{1-\cos(\pi \tau_e)}{2},
\end{equation}
where $\tau_e$ increases from 0 to 1 over the curriculum ramp. For an anchor with $N_i$
eligible neighbors, the window center is
\begin{equation}
w_i(e)=\lfloor(1-\eta(e))(N_i-1)\rfloor.
\end{equation}
As training progresses, batches shift from easier, semantically distant negatives toward
harder, semantically closer negatives. This makes the retrieval objective increasingly
sensitive to fine-grained interaction differences.

\section{Results}
\label{sec:results}

\subsection{Dataset} We evaluate MIME retrieval on the Inter-X \cite{xu2024inter} dataset, a large-scale dataset containing over 11k interactive motion sequences, 8 million frames, and 34k textual descriptions. For downstream and frozen MIME testing we use InterHuman \cite{liang2024intergen}, which has over 7k interactive motion sequences, 107 million frames and 24k descriptions.

\subsection{Implementation Details}

All models use a latent dimension of $512$, $L_c=4$ co-attention layers, $4$ attention heads, and a dropout rate of $0.1$. We train the models using AdamW~\cite{loshchilov2017decoupled} with a learning rate of $1\times10^{-4}$, weight decay of $1\times10^{-4}$, and a batch size of $128$. Motion sequences sampled at 30 fps are padded or truncated to a maximum length of $N=300$ frames. When curriculum learning is enabled, the sampling hardness is gradually increased over $E_{\mathrm{curr}}=10$ epochs after a $3$-epoch warm-up, up to a maximum hardness of $\eta_{\max}=0.25$. We adopt a $70/10/20$ train/validation/test split and retain a sufficiently large held-out test set for retrieval evaluation using galleries containing up to $2{,}000$ samples.

\subsection{Evaluation Metrics} Following the training objective in \cref{sec:methods}, we report text-to-motion retrieval, where a caption retrieves its paired motion, and motion-to-text retrieval, where a motion pair retrieves its paired caption. Performance is measured using recall at $K$ (R@$K$)~\cite{guo2022generating}, where retrieval is correct if the ground-truth pair appears in the top $K$ ranked candidates. Gallery size refers to the number of unseen samples from which the model selects. Unless otherwise stated, evaluation uses all available captions for each selected motion and motion-to-text recall treats any caption for the motion as correct. For each gallery size, we use the same deterministic subset across methods and seeds.

For downstream evaluation, MM Dist~\cite{guo2022generating} measures the average distance between matched text and motion embeddings, with lower values indicating better semantic alignment. FID~\cite{heusel2017gans} measures how closely the distribution of generated motions matches that of real motions, where lower is better. Diversity~\cite{guo2020action2motion} measures variation across generated motions. Downstream MM Distance and R-precision metrics are computed using the standard InterHuman evaluator employed by the original generation protocols.

\subsection{Retrieval Baselines}
We compare MIME against three encoder baselines adapted to the same Inter-X split, text captions, base 135-dimensional per-actor motion representation, text encoder, contrastive training objective, and retrieval evaluation protocol. For TMR Early Fusion, we concatenate the two base actor streams along the feature dimension before temporal encoding. For TMR Late 
Fusion, we encode the actors independently and fuse their global representations through a learned projection head. Both variants are trained with the same symmetric text-motion contrastive objective and evaluated using the same retrieval metrics as MIME. MIME differs by constructing the proposed frame-level interaction features and runtime relation stream as components of its interaction-aware architecture.

We also include LaMP as a controlled representation-backbone baseline. Rather than reproducing LaMP's full pretraining or generation pipeline, we use its motion representation backbone under the same early-fusion retrieval protocol: the two actor streams are concatenated at the input and mapped into the shared text-motion embedding space using the same retrieval loss. This isolates the effect of the motion encoder architecture while keeping the dataset split, text encoder, training objective, and evaluation protocol fixed across methods.

\begin{table*}[tb]
\centering
\caption{Retrieval performance comparison between TMR, LaMP, and MIME across retrieval gallery sizes. Results are reported as mean ± standard deviation across three training runs with different random seeds}
\label{tab:retrieval_results}
\small
\setlength{\tabcolsep}{6pt}
\renewcommand{\arraystretch}{1.15}
\resizebox{\textwidth}{!}{%
\begin{tabular}{llcccccccc}
\toprule
\multirow{2}{*}{\textbf{Gallery size}} &
\multirow{2}{*}{\textbf{Model}} &
\multicolumn{4}{c}{\textbf{Text-motion retrieval}} &
\multicolumn{4}{c}{\textbf{Motion-text retrieval}} \\
\cmidrule(lr){3-6}
\cmidrule(lr){7-10}
& & R@1$\uparrow$ & R@3$\uparrow$ & R@5$\uparrow$ & R@10$\uparrow$
& R@1$\uparrow$ & R@3$\uparrow$ & R@5$\uparrow$ & R@10$\uparrow$ \\
\midrule

\multirow{4}{*}{500}
& TMR Late Fusion
& $33.14 \pm 0.09$ & $53.00 \pm 0.85$ & $64.70 \pm 0.42$ & $76.07 \pm 0.37$
& $42.80 \pm 6.51$ & $62.00 \pm 3.68$ & $72.30 \pm 2.40$ & $83.80 \pm 1.41$ \\
& TMR Early Fusion
& $34.64 \pm 1.29$ & $55.35 \pm 1.10$ & $64.11 \pm 0.79$ & $75.42 \pm 0.28$
& $45.13 \pm 1.03$ & $66.60 \pm 1.78$ & $75.33 \pm 1.86$ & $85.53 \pm 0.76$ \\
& LaMP
& $30.70 \pm 1.97$ & $53.13 \pm 1.80$ & $65.14 \pm 2.07$ & $77.47 \pm 1.87$
& $37.20 \pm 1.80$ & $61.40 \pm 3.00$ & $71.70 \pm 2.10$ & $83.60 \pm 1.20$ \\
& MIME
& $\mathbf{38.87 \pm 0.13}$ & $\mathbf{62.20 \pm 0.87}$ & $\mathbf{71.57 \pm 1.44}$ & $\mathbf{82.47 \pm 0.34}$
& $\mathbf{47.10 \pm 1.30}$ & $\mathbf{71.50 \pm 0.30}$ & $\mathbf{79.30 \pm 0.30}$ & $\mathbf{90.40 \pm 0.00}$ \\
\midrule

\multirow{4}{*}{1000}
& TMR Late Fusion
& $22.93 \pm 0.71$ & $41.05 \pm 0.45$ & $51.02 \pm 0.16$ & $64.25 \pm 0.74$
& $30.65 \pm 2.05$ & $49.20 \pm 1.70$ & $59.25 \pm 2.19$ & $71.65 \pm 1.06$ \\
& TMR Early Fusion
& $24.83 \pm 1.12$ & $43.74 \pm 1.52$ & $52.52 \pm 1.14$ & $64.75 \pm 1.05$
& $31.27 \pm 1.65$ & $52.97 \pm 2.43$ & $63.10 \pm 2.44$ & $76.07 \pm 1.56$ \\
& LaMP
& $21.27 \pm 1.77$ & $40.45 \pm 1.85$ & $51.44 \pm 1.87$ & $65.07 \pm 2.00$
& $26.35 \pm 2.25$ & $47.45 \pm 4.15$ & $57.55 \pm 3.95$ & $71.70 \pm 1.80$ \\
& MIME
& $\mathbf{28.94 \pm 0.87}$ & $\mathbf{49.52 \pm 0.75}$ & $\mathbf{59.72 \pm 0.25}$ & $\mathbf{72.77 \pm 0.37}$
& $\mathbf{34.60 \pm 1.10}$ & $\mathbf{57.55 \pm 0.75}$ & $\mathbf{66.35 \pm 0.05}$ & $\mathbf{79.00 \pm 0.10}$ \\
\midrule

\multirow{4}{*}{2000}
& TMR Late Fusion
& $16.60 \pm 0.83$ & $31.89 \pm 0.91$ & $40.10 \pm 0.49$ & $52.94 \pm 0.50$
& $22.23 \pm 0.53$ & $39.25 \pm 2.47$ & $47.90 \pm 2.40$ & $61.20 \pm 1.84$ \\
& TMR Early Fusion
& $18.32 \pm 1.17$ & $33.31 \pm 1.45$ & $41.90 \pm 1.66$ & $53.53 \pm 1.51$
& $24.20 \pm 0.69$ & $42.42 \pm 1.28$ & $52.20 \pm 1.69$ & $64.90 \pm 1.35$ \\
& LaMP
& $14.57 \pm 1.10$ & $29.26 \pm 1.75$ & $38.37 \pm 1.85$ & $51.91 \pm 2.02$
& $17.70 \pm 1.75$ & $35.00 \pm 3.05$ & $44.40 \pm 3.95$ & $58.00 \pm 3.30$ \\
& MIME
& $\mathbf{20.67 \pm 0.68}$ & $\mathbf{38.23 \pm 1.44}$ & $\mathbf{47.74 \pm 0.88}$ & $\mathbf{60.72 \pm 1.26}$
& $\mathbf{25.78 \pm 0.23}$ & $\mathbf{44.90 \pm 0.85}$ & $\mathbf{55.58 \pm 0.98}$ & $\mathbf{68.85 \pm 0.70}$ \\
\bottomrule
\end{tabular}%
}
\end{table*}

\subsection{Quantitative Evaluation}

As shown in \cref{tab:retrieval_results}, MIME consistently outperforms both early and late fusion baselines across all categories. At a gallery size of 500, MIME achieves a text-to-motion R@1 of 38.87, compared to 33.14 for TMR Late Fusion, 34.64 for early-fusion TMR, and 30.7 for LaMP. For motion to text retrieval, MIME improves over the strongest baseline at R@1, R@3, and R@5 by 4.3\%, 7.4\%, and 5.3\% respectively.

The advantage of MIME becomes more pronounced at larger gallery sizes. At the more difficult 2,000-example gallery, MIME improves text-to-motion R@1 to 20.67, achieving a 12.8\% relative gain (+2.35 absolute points) over the TMR Early Fusion baseline (18.32), while maintaining even larger margins over late-fusion TMR (16.60). MIME similarly improves R@5 to 47.74, compared to TMR Early Fusion's 41.90. MIME also outperforms LaMP at this scale, improving text-to-motion R@1 from 14.57 to 20.67 and motion-to-text R@1 from 17.70 to 25.78. These results suggest that MIME learns a more discriminative latent space for fine-grained interactive motion.

\begin{figure*}[tb]
    \centering
    \includegraphics[
        width=\textwidth,
        keepaspectratio
    ]{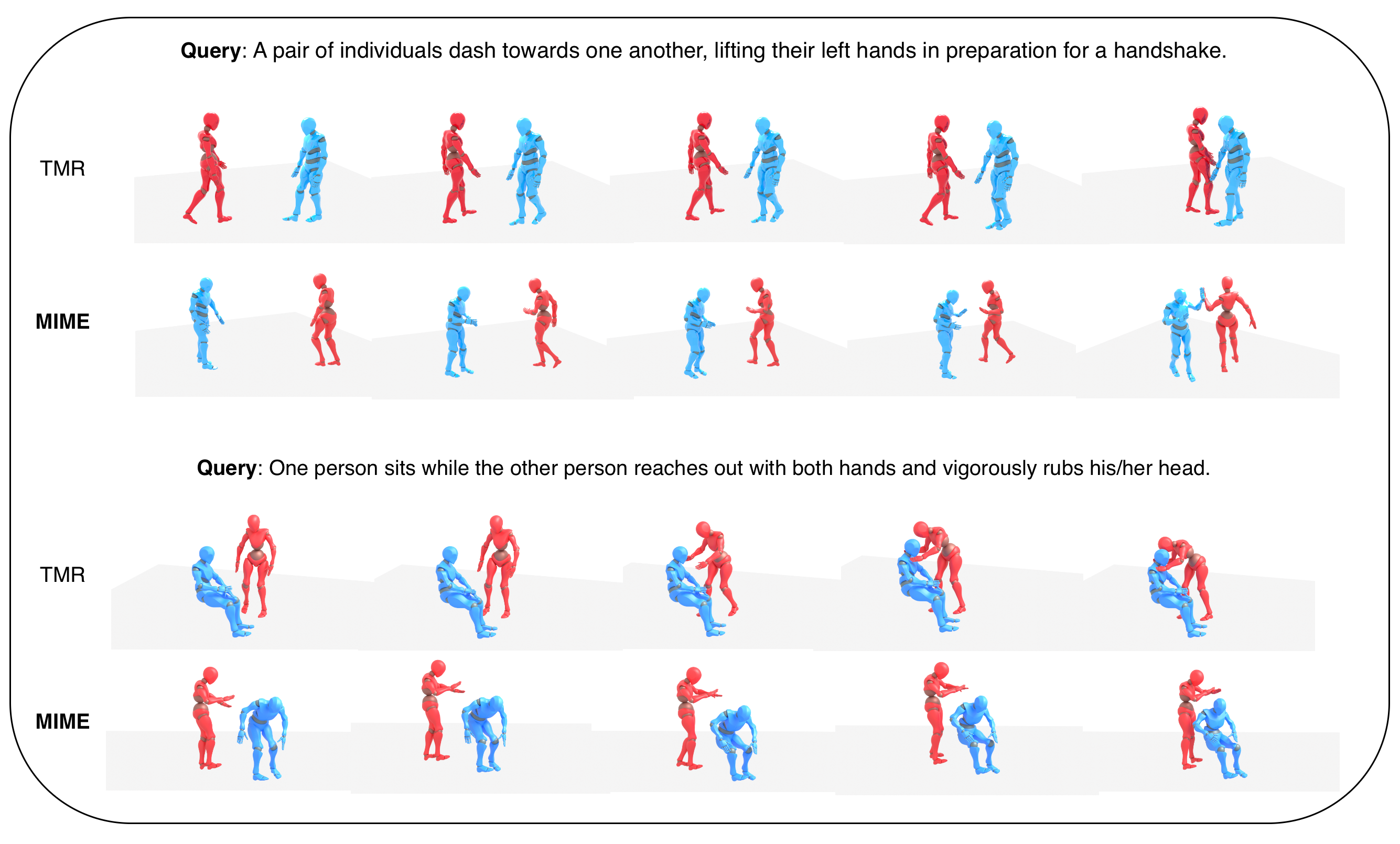}
    \caption{
     Qualitative text-to-motion retrieval comparison between MIME and TMR Early Fusion. Each row shows the top-1 motion retrieved for the corresponding text query, visualized at five representative frames. In both examples, MIME retrieves the paired ground-truth motion, whereas TMR Early Fusion retrieves a different motion sequence. In the top example TMR fails to retrieve a dash or hand lift, and in the bottom sequence TMR fails to retrieve the head rub. }
    \label{fig:qualitative_retrieval}
\end{figure*}

\subsection{Qualitative Evaluation}
\begin{figure}[t]
    \centering
    \includegraphics[width=\linewidth]{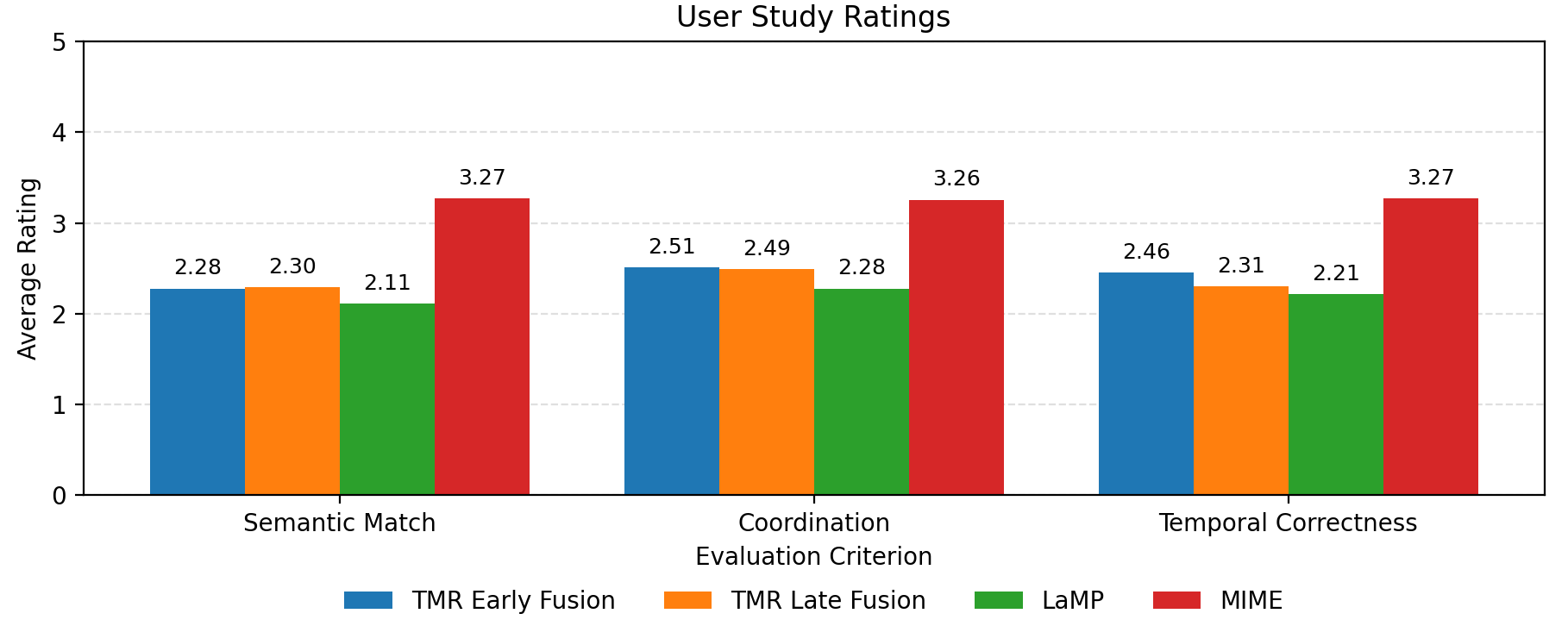}
    \caption{
    User study comparison between MIME and baselines. Participants rated retrieved motions on semantic alignment, interaction alignment, and temporal correctness using a 5-point Likert scale. Higher scores indicate better rated query-motion alignment.
    }
    \label{fig:user_study_results}
\end{figure}

We conduct a user study to qualitatively evaluate the text to motion retrieval performance of MIME vs. the baselines as seen in Fig \ref{fig:user_study_results}. On the Likert scale users scored MIME retrieved motions 42\% more aligned, 30\% more coordinated, and 32\% more temporally correct than the top baseline. Please refer to the Supplementary for more details on the User Study.

 \cref{fig:qualitative_retrieval} presents two representative text-to-motion retrieval examples comparing MIME R@1 with early-fusion TMR R@1. In the first example, the query describes two individuals dashing toward one another while raising their left hands in preparation for a handshake. MIME retrieves the paired ground-truth motion, capturing both the rapid approach and the raised-hand gesture. In contrast, TMR retrieves a sequence in which the actors approach more slowly and do not exhibit the requested hand motion. In the second example, the query describes one person sitting while the other rubs their head with both hands. MIME again retrieves the paired ground-truth sequence and preserves the head-directed interaction. TMR captures the broader configuration of one seated and one standing actor, but retrieves an interaction resembling an upper-back pat rather than head rubbing. These examples qualitatively suggest that MIME better distinguishes fine-grained relational actions, whereas early fusion can preserve the overall scene while missing the specific interaction described by the text.

\begin{figure}[tb]
    \centering
    \includegraphics[width=\linewidth]{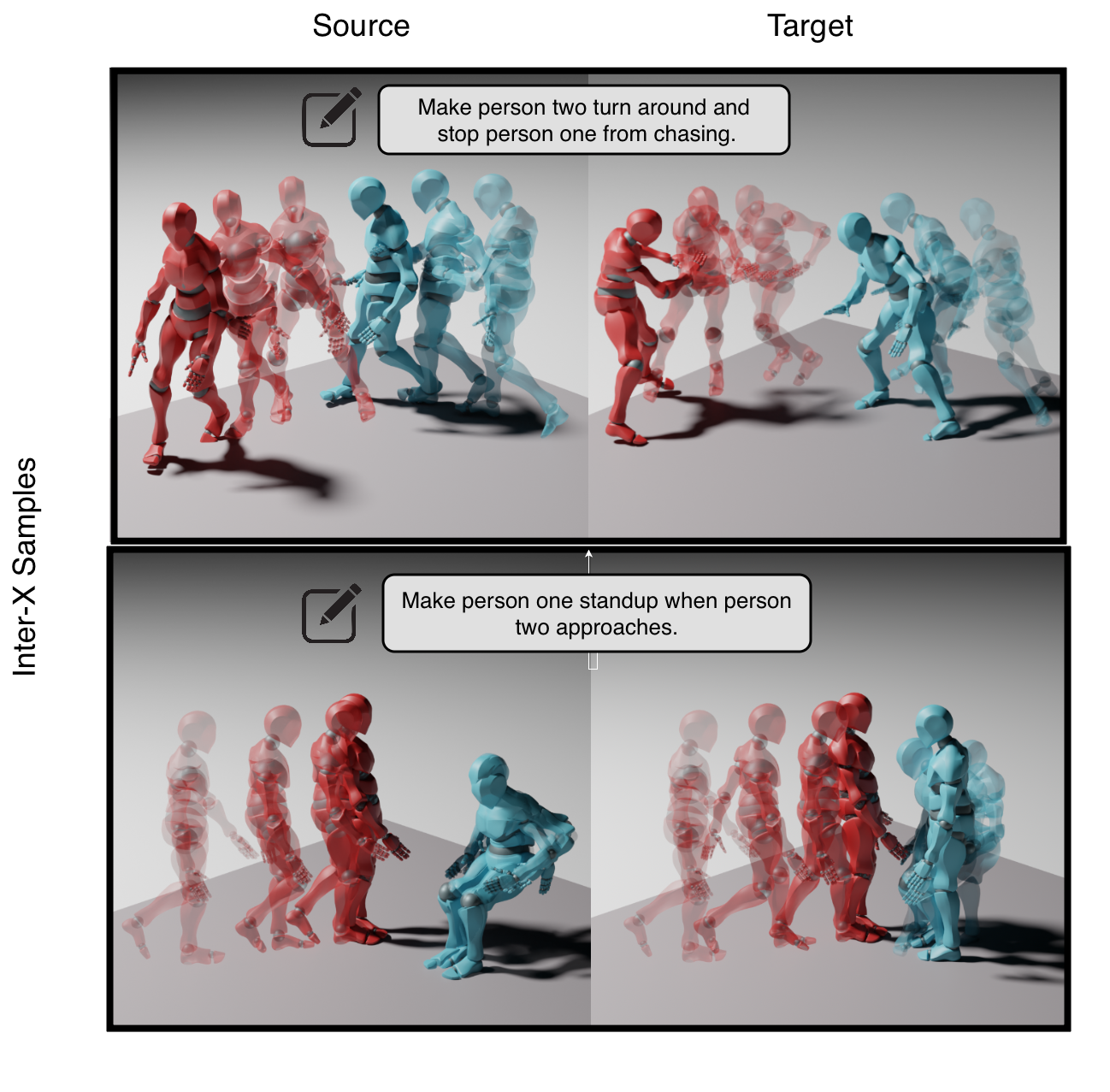}
    \caption{Qualitative retrieval results framed in a downstream editing scenario. MIME retrieves semantically related interaction pairs that can provide useful source-target examples for interactive motion editing.}
    \label{fig:qualitative_editing}
\end{figure}

We also examine retrieval in an editing-oriented setting. Constructing motion-editing pairs requires examples that are closely aligned in pose and semantics while differing in a small, meaningful way. As shown in \cref{fig:qualitative_editing}, MIME retrieves highly similar interactions with localized differences, such as one person standing up in one clip but remaining seated in the other. These pairs provide useful source-target examples for retrieval-based dataset construction and editing supervision. Motions are rendered in Blender, with specific frames selected for visual clarity.

\begin{table}[tb]
\centering
\caption{Ablation study of MIME components across retrieval gallery sizes. Results are mean $\pm$ standard deviation across three training runs.}
\label{tab:mime_ablation}
\scriptsize
\setlength{\tabcolsep}{2.0pt}
\renewcommand{\arraystretch}{1.08}

\resizebox{\columnwidth}{!}{%
\begin{tabular}{llcccccc}
\toprule
\multirow{2}{*}{\textbf{Size}} & \multirow{2}{*}{\textbf{Model}}
& \multicolumn{3}{c}{\textbf{T2M}}
& \multicolumn{3}{c}{\textbf{M2T}} \\
\cmidrule(lr){3-5} \cmidrule(lr){6-8}
& & R@1 & R@5 & R@10 & R@1 & R@5 & R@10 \\
\midrule

\multirow{4}{*}{500}
& Full
& $38.87{\pm}0.13$
& $71.57{\pm}1.44$
& $\mathbf{82.47{\pm}0.34}$
& $47.10{\pm}1.30$
& $79.30{\pm}0.30$
& $\mathbf{90.40{\pm}0.00}$ \\

& --IF
& $37.64{\pm}0.64$
& $69.10{\pm}1.77$
& $80.00{\pm}0.67$
& $45.90{\pm}0.90$
& $76.80{\pm}0.80$
& $88.30{\pm}0.50$ \\

& --Cur
& $\mathbf{39.16{\pm}0.23}$
& $\mathbf{72.23{\pm}1.30}$
& $81.44{\pm}0.44$
& $\mathbf{48.70{\pm}0.90}$
& $\mathbf{80.00{\pm}0.80}$
& $88.70{\pm}0.90$ \\

& --CA
& $37.27{\pm}0.94$
& $68.77{\pm}0.36$
& $80.30{\pm}0.50$
& $45.60{\pm}0.40$
& $79.40{\pm}1.60$
& $88.80{\pm}0.60$ \\
\midrule

\multirow{4}{*}{1000}
& Full
& $\mathbf{28.94{\pm}0.87}$
& $\mathbf{59.72{\pm}0.25}$
& $\mathbf{72.77{\pm}0.37}$
& $34.60{\pm}1.10$
& $66.35{\pm}0.05$
& $79.00{\pm}0.10$ \\

& --IF
& $27.31{\pm}0.99$
& $57.34{\pm}0.13$
& $69.90{\pm}0.50$
& $33.40{\pm}0.00$
& $65.20{\pm}0.20$
& $77.80{\pm}1.00$ \\

& --Cur
& $28.48{\pm}0.59$
& $59.22{\pm}0.01$
& $72.14{\pm}1.27$
& $\mathbf{36.15{\pm}1.35}$
& $\mathbf{68.00{\pm}0.60}$
& $\mathbf{79.65{\pm}0.05}$ \\

& --CA
& $27.35{\pm}0.22$
& $56.63{\pm}0.50$
& $68.80{\pm}0.70$
& $35.65{\pm}0.15$
& $66.15{\pm}0.35$
& $77.40{\pm}0.30$ \\
\midrule

\multirow{4}{*}{2000}
& Full
& $\mathbf{20.67{\pm}0.68}$
& $\mathbf{47.74{\pm}0.88}$
& $\mathbf{60.72{\pm}1.26}$
& $25.78{\pm}0.23$
& $\mathbf{55.58{\pm}0.98}$
& $\mathbf{68.85{\pm}0.70}$ \\

& --IF
& $19.79{\pm}0.77$
& $46.15{\pm}0.36$
& $58.79{\pm}0.21$
& $24.80{\pm}0.40$
& $54.00{\pm}0.55$
& $67.72{\pm}0.43$ \\

& --Cur
& $20.31{\pm}0.47$
& $47.38{\pm}0.02$
& $60.54{\pm}0.01$
& $\mathbf{26.05{\pm}1.40}$
& $55.00{\pm}1.80$
& $68.12{\pm}0.67$ \\

& --CA
& $19.48{\pm}0.14$
& $45.17{\pm}0.32$
& $58.00{\pm}0.45$
& $25.68{\pm}0.38$
& $54.55{\pm}1.25$
& $67.32{\pm}0.02$ \\
\bottomrule
\end{tabular}%
}

\vspace{2pt}
\footnotesize{
--IF: w/o interaction features; --Cur: w/o curriculum; --CA: w/o co-attention.
}
\end{table}

\subsection{Ablation Study}
We report ablations in \cref{tab:mime_ablation} to isolate the contribution of MIME's main components. At the 2,000-sample gallery, the full model leads in five of the six reported metrics. The no-curriculum variant is slightly higher on motion-to-text R@1, while the full model performs better on all text-to-motion metrics and on motion-to-text R@5 and R@10. At smaller gallery sizes, several ablated variants occasionally outperform the full model, suggesting that the benefits of the complete architecture become more apparent as retrieval becomes more difficult.

The most consistent contribution comes from the explicit interaction features. Removing them reduces performance across all gallery sizes and lowers text-to-motion retrieval at a gallery size of 2,000 from 20.67 to 19.79 R@1, from 47.74 to 46.15 R@5, and from 60.72 to 58.79 R@10. This indicates that relative motion cues provide useful information beyond the base features and streams.

Co-attention and curriculum sampling show less uniform gains at the 500- and 1,000-sample galleries, but both contribute at the more challenging 2,000-sample setting. Removing co-attention weakens MIME's ability to model relationships between the two actor streams, while removing the curriculum reduces its ability to distinguish among closely related interactions. Overall, the ablations indicate that explicit interaction features provide the most consistent improvement, while co-attention and curriculum sampling are most beneficial under harder retrieval conditions.

\begin{table*}[t]
\centering
\caption{Secondary downstream validation on InterHuman via a semantic-conditioning probe. We integrate MIME as a frozen auxiliary prior into TIMotion and InterMask~\cite{javed2024intermask}. Generated-model results are reported as mean $\pm$ 95\% confidence interval across nine evaluation runs; MIME is projected into the conditioning latent space through a lightweight adapter. Bolding indicates the strongest generated-model result for the corresponding metric. $\downarrow$ indicates lower is better and $\uparrow$ indicates higher is better.}
\label{tab:timotion_downstream}
\small
\setlength{\tabcolsep}{4pt}
\renewcommand{\arraystretch}{1.10}
\resizebox{\textwidth}{!}{
\begin{tabular}{lcccccc}
\toprule
\textbf{Model} &
\textbf{MM Dist.}$\downarrow$ &
\textbf{R@1}$\uparrow$ &
\textbf{R@2}$\uparrow$ &
\textbf{R@3}$\uparrow$ &
\textbf{FID}$\downarrow$ &
\textbf{Diversity}\\
\midrule

Ground Truth
& $3.7850 \pm 0.0018$
& $0.4242 \pm 0.0079$
& $0.6038 \pm 0.0051$
& $0.7045 \pm 0.0060$
& $0.2897 \pm 0.0139$
& $7.7604 \pm 0.0856$\\

TIMotion
& $3.7805 \pm 0.0019$
& $0.4732 \pm 0.0141$
& $0.6312 \pm 0.0120$
& $0.7094 \pm 0.0103$
& $5.5754 \pm 0.2427$
& $7.9598 \pm 0.0643$ \\

TIMotion + MIME Prior
& $\mathbf{3.7758 \pm 0.0012}$
& $\mathbf{0.4924 \pm 0.0107}$
& $\mathbf{0.6404 \pm 0.0119}$
& $\mathbf{0.7115 \pm 0.0128}$
& $5.5645 \pm 0.0229$
& $7.9120 \pm 0.0929$ \\

InterMask
& $3.7910 \pm 0.0015$
& $0.4372 \pm 0.0117$
& $0.5925 \pm 0.0043$
& $0.6793 \pm 0.0058$
& $\mathbf{5.0859 \pm 0.2504}$
& $8.0609 \pm 0.0858$ \\

InterMask + MIME
& $3.7902 \pm 0.0014$
& $0.4489 \pm 0.0117$
& $0.6010 \pm 0.0105$
& $0.6803 \pm 0.0085$
& $5.3194 \pm 0.1084$
& $7.9518 \pm 0.1358$ \\

\bottomrule
\end{tabular}
}
\end{table*}

\subsection{Downstream Task Validation}

As a secondary downstream validation, we test whether MIME's learned representation transfers beyond its primary retrieval setting. We integrate MIME as a frozen auxiliary prior into TIMotion and InterMask and evaluate both models on InterHuman, a dataset entirely unseen during MIME's contrastive training. MIME's interaction-aware text representation is projected through a lightweight trainable adapter and fused with each generator's original conditioning signal. The adapter is a small residual MLP with a Linear--GELU--Linear projection. Thus, the experiment evaluates cross-dataset transfer of the frozen MIME representation rather than adaptation of MIME to the downstream distribution.

As shown in \cref{tab:timotion_downstream}, adding MIME to TIMotion improves retrieval-based semantic alignment. R@1 increases from $0.4732$ to $0.4924$, and R@2 increases from $0.6312$ to $0.6404$, while the change at R@3 is smaller, from $0.7094$ to $0.7115$. Mean FID remains nearly unchanged, moving from $5.5754$ to $5.5645$.

FID measures distribution-level similarity between generated and real motions, whereas retrieval-based metrics measure how closely a generated motion corresponds to its conditioning description. The TIMotion results indicate that MIME provides a more semantically informative conditioning signal while retaining comparable distribution-level fidelity. This property is useful for text-guided animation and motion editing, where generated motion should be both plausible and consistent with the requested interaction.

A similar alignment trend appears with InterMask. Adding MIME increases R@1 from $0.4372$ to $0.4489$ and R@2 from $0.5925$ to $0.6010$, while the change at R@3 is minimal. InterMask exhibits a modest FID increase from $5.0859$ to $5.3194$, indicating a trade-off between the improved retrieval-based alignment and distribution-level fidelity for this generator. InterMask's discrete tokenization may make its predictions more sensitive to changes in the conditioning signal, although the present experiments do not isolate the cause of this trade-off.

Overall, the results show initial evidence that interaction semantics learned on Inter-X transfer through a frozen MIME representation to the unseen InterHuman dataset. This supports MIME's broader utility as a semantic representation for conditioning interactive motion generation.

\section{Conclusion}
\label{sec:conclusion}

We introduced MIME, an interaction-aware multimodal encoder that aligns language with two-person motion while preserving actor-specific streams. MIME models frame-synchronous relationships through explicit interaction features and bidirectional co-attention. On Inter-X, MIME consistently outperforms early- and late-fusion baselines, with its strongest gains appearing at the challenging 2,000-sample gallery. As a frozen auxiliary prior on the unseen InterHuman dataset, MIME also improves retrieval-based semantic alignment in TIMotion and InterMask while maintaining comparable FID in TIMotion. These results demonstrate the value of explicitly modeling multi-person relationships and support MIME as a reusable representation for interactive-motion retrieval and conditioning.

\paragraph{Limitations \& Future Work}
MIME is currently limited to two-person interactions, and its downstream results demonstrate improved semantic alignment rather than generative fidelity. Future work will extend MIME to variable-size groups, richer contact and synchronization cues, and tighter integration with generative models.

{
    \small
    \bibliographystyle{ieeenat_fullname}
    \bibliography{main}
}

\clearpage
\maketitlesupplementary

\renewcommand{\thesection}{\Alph{section}}
\setcounter{section}{0}

\section{Reproducibility}
Full code for this project along with the trained checkpoints will be made open source and publicly available upon paper acceptance.

\section{Application to Editing}

\begin{figure}[tb]
    \centering
    \includegraphics[width=\linewidth]{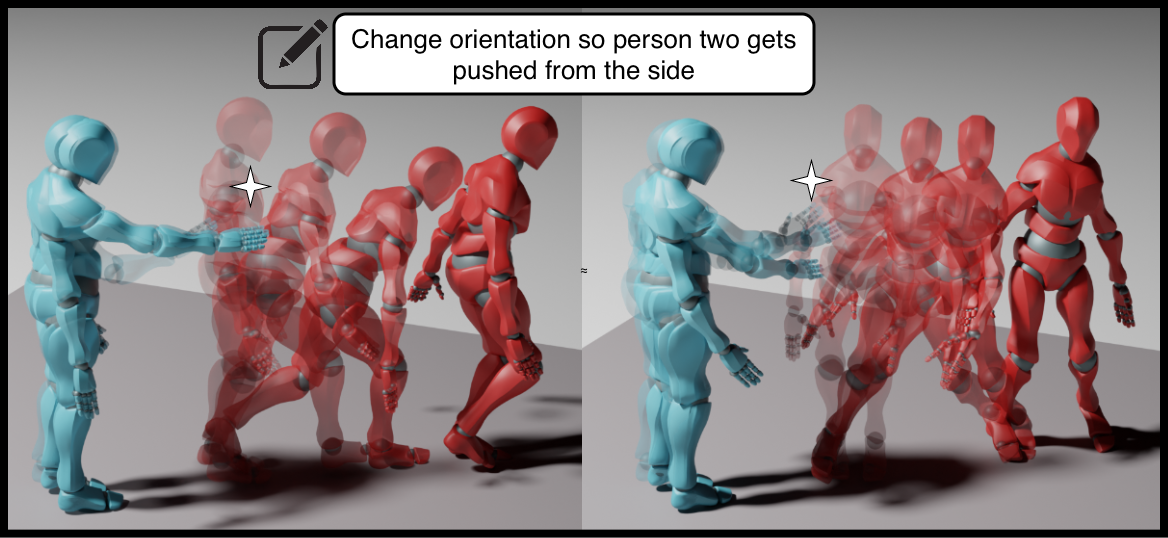}
    \caption{Example of an edit pair retrieved using MIME. First motion is ground truth, second motion is R@5 in Inter-X dataset. They are closely aligned, but have a difference which can be utilized for editing. }
    \label{fig:qualitative_editing_sup}
\end{figure}

\noindent\textbf{Editing-oriented retrieval and pair construction.}
Learned representations have previously been used to mine source--target pairs for editing supervision. MotionFix~\cite{athanasiou2024motionfix} retrieves semantically similar single-person motions using TMR \cite{petrovich2023tmr} embeddings before collecting natural-language descriptions of their differences. More recently, InterEdit3D~\citesupp{yang2026interedit} applies window-level retrieval with a frozen TMR encoder to construct editing triplets for two-person motion. Related pair-mining strategies have also been explored for static pose correction~\citesupp{delmas2023posefix} and language-guided 3D shape editing~\citesupp{Achlioptas_2023_CVPR}.

These works establish embedding-space retrieval as a practical mechanism for finding examples that preserve substantial source content while differing in a concise, describable attribute. MIME offers a complementary representation for this setting because it encodes both actor-specific motion and frame-synchronous interaction structure. As illustrated in Fig.~\ref{fig:qualitative_editing_sup}, neighboring motions retrieved by MIME preserve the broad pushing interaction while differing in the response trajectory of the second actor. Such results could serve as candidate pairs for instructions that alter the direction or magnitude of the reaction while retaining the underlying interaction.

\section{Additional Experiments \& Analysis}

\begin{figure}[tb]
    \centering
    \includegraphics[width=\linewidth]{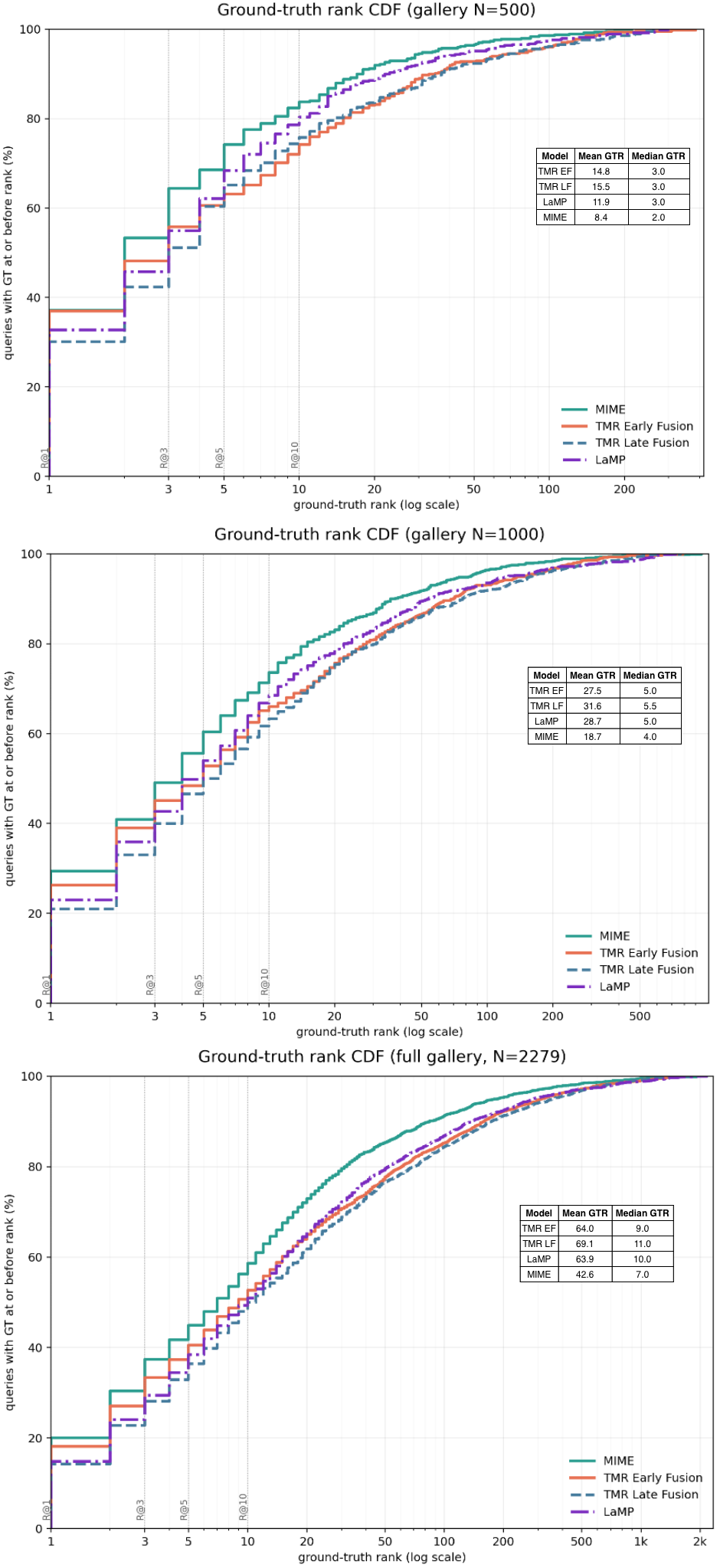}
    
    \caption{Ground-truth rank distributions for text-to-motion retrieval with gallery sizes of 500, 1,000, and the full 2,279-sample test set. Each curve shows the percentage of queries whose paired motion is retrieved at or before a given rank; curves that rise earlier indicate better retrieval. MIME consistently shifts the distribution toward lower ranks and achieves the best mean and median ground-truth rank at every gallery size}
    \label{fig:CDF_Supp}
\end{figure}

\noindent\textbf{Ground-truth rank distribution.}
Beyond evaluating retrieval at individual recall thresholds, Fig.~\ref{fig:CDF_Supp} characterizes the complete distribution of ground-truth ranks across three gallery sizes. MIME's CDF consistently rises earlier than those of the baselines, indicating that a larger fraction of queries retrieve their paired motion within any given rank threshold. Relative to the strongest baseline, MIME reduces the mean ground-truth rank by 29.4\%, 32.0\%, and 33.3\% for gallery sizes of 500, 1,000, and 2,279, respectively. MIME also achieves lower median ranks of 2, 4, and 7, compared with 3, 5, and 9 for the strongest baselines. These results show that MIME's advantage is not limited to selected R@$K$ thresholds, but extends across the full rank distribution, including more difficult retrieval cases.\\

\begin{figure}[tb]
    \centering
    \includegraphics[width=\linewidth]{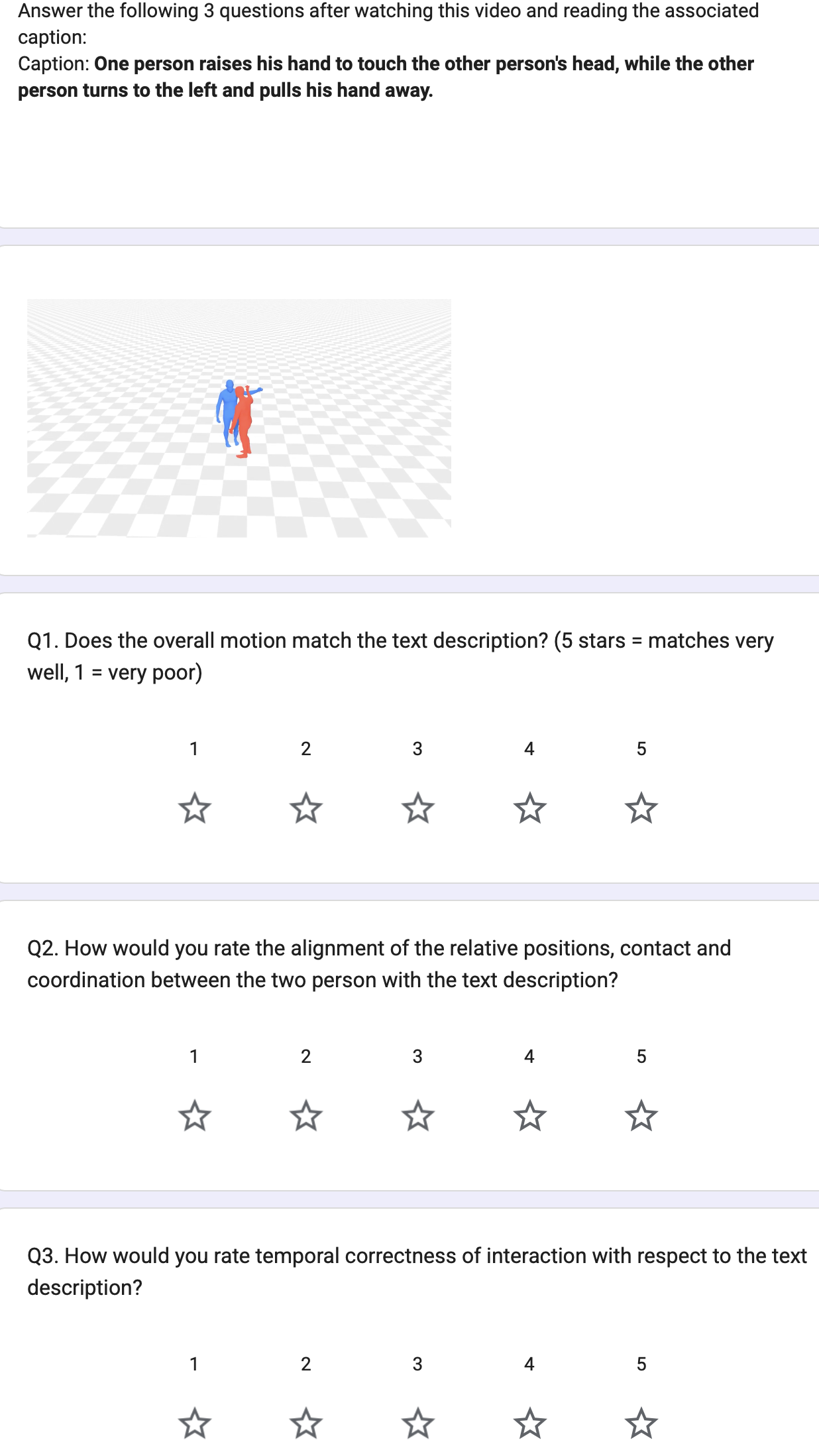}
    \caption{Screenshot of the user study using Google Forms. Each question had the attached video + questions to rate}
    \label{fig:user_study_ui}
\end{figure}

\noindent\textbf{User Study.}
Our user study evaluates whether motions retrieved by MIME are perceived as more semantically aligned with input interaction descriptions than those retrieved by the baselines. We sampled 10 text queries from the held-out Inter-X test split and obtained the top-1 retrieved motion from TMR Early Fusion, TMR Late Fusion, LaMP and MIME, matching the methods used in our quantitative retrieval evaluation. For each query, all retrieved motions were rendered using the same camera viewpoint, skeleton style, frame rate, and duration. Method identities were hidden from participants, and the display order was randomized independently for each query to avoid positional bias. We recruited 30 university students and graduates. In each trial, participants were shown a series of text descriptions together with the retrieved motions and asked to rate each result on a 5-point Likert scale for three criteria: \textit{(1) How well does the motion align semantically with the text description? (2) How would you rate the alignment of the relative positions, contact and coordination between the two people considering the text? (3) How would you rate temporal correctness of interaction with respect to the text description?} Fig.~\ref{fig:user_study_ui} shows the user study form in our evaluation. We compute the mean score for each method across all participants and queries, and report the comparison in Fig.~\ref{fig:user_study_results}. The results show that participants consistently rate motions retrieved by MIME higher than those retrieved by the baselines across all evaluation criteria. The strong MIME performance complements the quantitative results, especially considering that these queries were all done with a 1k gallery test set where MIME reduces mean ground-truth rank by 32\%.

\noindent\textbf{Efficiency Analysis}

\begin{table}[t]
    \centering
    \caption{
    Computational efficiency comparison. Parameter counts are reported as
    total and motion-side parameters, respectively; all models use the same
    151.3M-parameter text encoder. Motion-encoding latency excludes text
    encoding. Total training time denotes the complete wall-clock duration
    of each run.
    }
    \label{tab:efficiency}
    \resizebox{\linewidth}{!}{%
    \begin{tabular}{lcccc}
        \toprule
        \textbf{Model}
        & \textbf{Params.}
        & \textbf{Time/Epoch}
        & \textbf{Total Train Time}
        & \textbf{Motion Latency} \\
        & \textbf{Total / Motion (M)}
        & \textbf{(s)}
        & \textbf{(h:mm:ss)}
        & \textbf{(ms/sample)} \\
        \midrule

        TMR Early Fusion
        & 158.0 / 6.7
        & 54.7
        & 0:46:35
        & 0.404 \\

        TMR Late Fusion
        & 165.4 / 14.1
        & 73.3
        & 1:06:56
        & 0.678 \\

        LaMP
        & 162.3 / 11.0
        & 30.0
        & 0:51:41
        & 0.191 \\

        MIME
        & 177.7 / 26.4
        & 95.8
        & 1:47:18
        & 1.477 \\

        \midrule

        MIME 3L
        & 171.4 / 20.1
        & 62.3
        & 0:54:55
        & 1.22 \\

        MIME 2L
        & 165.0 / 13.8
        & 58.5
        & 0:42:41
        & 0.86 \\

        MIME 2L 1FF
        & 162.9 / 11.7
        & 57.3
        & 0:40:04
        & 0.80 \\

        \bottomrule
    \end{tabular}%
    }
\end{table}

\begin{table*}[tb]
\centering
\caption{Comparison of MIME architectural variants across retrieval gallery sizes.}
\label{tab:mime_depth_ablation}
\scriptsize
\setlength{\tabcolsep}{2.0pt}
\renewcommand{\arraystretch}{1.08}

\resizebox{\textwidth}{!}{%
\begin{tabular}{lllcccccc}
\toprule
\multirow{2}{*}{\textbf{Size}}
& \multirow{2}{*}{\textbf{Model}}
& \multirow{2}{*}{\textbf{Params. (M)}}
& \multicolumn{3}{c}{\textbf{T2M}}
& \multicolumn{3}{c}{\textbf{M2T}} \\
\cmidrule(lr){4-6}
\cmidrule(lr){7-9}
& & & R@1 & R@5 & R@10 & R@1 & R@5 & R@10 \\
\midrule

\multirow{4}{*}{500}
& MIME 3L
& 20.08
& $37.87 \pm 0.00$
& $70.94 \pm 0.07$
& $81.23 \pm 0.90$
& $\mathbf{47.20 \pm 1.20}$
& $79.00 \pm 1.40$
& $88.20 \pm 1.20$ \\

& MIME 2L
& 13.77
& $36.91 \pm 0.17$
& $70.58 \pm 0.69$
& $\mathbf{82.67 \pm 0.55}$
& $44.20 \pm 1.13$
& $78.80 \pm 1.77$
& $88.20 \pm 1.66$ \\

& MIME 2L 1FF
& 11.67
& $36.96 \pm 1.01$
& $70.18 \pm 1.07$
& $81.64 \pm 0.79$
& $46.47 \pm 2.17$
& $78.07 \pm 1.60$
& $87.20 \pm 0.28$ \\

& MIME
& 26.40
& $\mathbf{38.87 \pm 0.13}$
& $\mathbf{71.57 \pm 1.44}$
& $82.47 \pm 0.34$
& $47.10 \pm 1.30$
& $\mathbf{79.30 \pm 0.30}$
& $\mathbf{90.40 \pm 0.00}$ \\
\midrule

\multirow{4}{*}{1000}
& MIME 3L
& 20.08
& $27.24 \pm 0.44$
& $58.05 \pm 0.12$
& $71.37 \pm 0.30$
& $34.30 \pm 0.60$
& $65.90 \pm 0.10$
& $\mathbf{79.35 \pm 1.55}$ \\

& MIME 2L
& 13.77
& $26.45 \pm 0.70$
& $58.22 \pm 0.29$
& $71.57 \pm 0.52$
& $33.00 \pm 0.49$
& $65.00 \pm 0.41$
& $77.90 \pm 0.28$ \\

& MIME 2L 1FF
& 11.67
& $25.69 \pm 0.91$
& $56.84 \pm 1.11$
& $71.11 \pm 0.83$
& $33.03 \pm 2.50$
& $\mathbf{66.50 \pm 1.84}$
& $79.27 \pm 2.17$ \\

& MIME
& 26.40
& $\mathbf{28.94 \pm 0.87}$
& $\mathbf{59.72 \pm 0.25}$
& $\mathbf{72.77 \pm 0.37}$
& $\mathbf{34.60 \pm 1.10}$
& $66.35 \pm 0.05$
& $79.00 \pm 0.10$ \\
\midrule

\multirow{4}{*}{2000}
& MIME 3L
& 20.08
& $19.39 \pm 0.13$
& $46.28 \pm 0.74$
& $59.05 \pm 0.49$
& $24.50 \pm 0.85$
& $54.10 \pm 0.95$
& $67.75 \pm 1.00$ \\

& MIME 2L
& 13.77
& $18.66 \pm 0.45$
& $45.52 \pm 0.45$
& $59.34 \pm 0.45$
& $23.70 \pm 0.36$
& $53.87 \pm 0.97$
& $68.02 \pm 0.45$ \\

& MIME 2L 1FF
& 11.67
& $18.73 \pm 1.24$
& $45.39 \pm 1.45$
& $58.86 \pm 1.34$
& $23.25 \pm 1.84$
& $53.58 \pm 2.36$
& $68.32 \pm 1.89$ \\

& MIME
& 26.40
& $\mathbf{20.67 \pm 0.68}$
& $\mathbf{47.74 \pm 0.88}$
& $\mathbf{60.72 \pm 1.26}$
& $\mathbf{25.78 \pm 0.23}$
& $\mathbf{55.58 \pm 0.98}$
& $\mathbf{68.85 \pm 0.70}$ \\

\bottomrule
\end{tabular}%
}
\end{table*}

MIME incurs greater computational cost than the baselines because its interaction-aware architecture maintains separate actor streams, constructs explicit frame-level relational features, and applies bidirectional co-attention. As shown in Table~\ref{tab:efficiency}, however, the absolute computational cost remains practical: the full model trains in approximately 1~h 47~min and encodes each motion sample in 1.48~ms. MIME also contains more motion-side parameters than the baselines. Because all methods share the same 151.3M-parameter text encoder, however, the difference in total model size is smaller, with 177.7M parameters for MIME compared with 158.0--165.4M for the baselines.

To assess whether MIME's retrieval gains are primarily explained by this additional capacity, we evaluate three compact variants: MIME 3L, which uses three co-attention layers; MIME 2L, which uses two; and MIME 2L 1FF, which additionally restricts the feed-forward width in each co-attention block to the 512-dimensional latent width. Tables~\ref{tab:efficiency} and~\ref{tab:mime_depth_ablation} show that MIME retains strong retrieval performance as its motion-side parameter count is reduced from 26.4M to 11.7M. In particular, MIME 2L 1FF uses 11.7M motion-side parameters---fewer than the 14.1M parameters of TMR Late Fusion---yet outperforms 16/18 baseline results across all reported gallery sizes and retrieval metrics. These results indicate that MIME's improvements are not solely attributable to increased parameter count, but instead reflect the effectiveness of its interaction-aware architectural design. The other variants also reduce computational cost while keeping results even closer to full MIME; All measurements were obtained on a single NVIDIA A100 40GB GPU.

\section{Retrieval under Semantically Similar Distractors}
\label{sec:hard_retrieval}

To test retrieval under more confusable candidate sets, we construct
query-specific hard galleries using a frozen CLIP ViT-B/32 text encoder that
is independent of all evaluated models. For each test caption, we compute its
CLIP text embedding and rank candidate motions according to the cosine
similarity between the query embedding and the CLIP embedding of each
candidate motion's first associated caption. We then form hard galleries of
size $g \in \{32,64,128\}$. Every test query is evaluated, and the same
queries and candidate galleries are used for all models.

\begin{table}[tb]
\centering
\caption{Text-to-motion retrieval under semantically similar distractors.
Query-specific hard galleries are constructed using frozen CLIP text
embeddings, with identical queries and candidates used for all models.
Results are reported from one checkpoint per model.}
\label{tab:hard_semantic_retrieval}

\scriptsize
\setlength{\tabcolsep}{2.0pt}
\renewcommand{\arraystretch}{1.08}

\resizebox{\columnwidth}{!}{%
\begin{tabular}{llcccc}
\toprule
\multirow{2}{*}{\textbf{Size}}
& \multirow{2}{*}{\textbf{Model}}
& \multicolumn{4}{c}{\textbf{T2M}} \\
\cmidrule(lr){3-6}
& & R@1 $\uparrow$
  & R@5 $\uparrow$
  & R@10 $\uparrow$
  & Mean Rank $\downarrow$ \\
\midrule

\multirow{4}{*}{32}
& MIME
& $\mathbf{46.40}$
& $\mathbf{85.61}$
& $\mathbf{93.86}$
& $\mathbf{3.18}$ \\
& TMR Early Fusion
& $44.21$
& $80.61$
& $91.67$
& $3.77$ \\
& TMR Late Fusion
& $40.88$
& $78.42$
& $92.19$
& $3.81$ \\
& LaMP
& $42.89$
& $81.40$
& $92.72$
& $3.60$ \\
\midrule

\multirow{4}{*}{64}
& MIME
& $\mathbf{40.44}$
& $\mathbf{79.12}$
& $\mathbf{90.09}$
& $\mathbf{4.47}$ \\
& TMR Early Fusion
& $38.33$
& $73.33$
& $84.21$
& $5.51$ \\
& TMR Late Fusion
& $35.18$
& $71.49$
& $84.30$
& $5.53$ \\
& LaMP
& $36.67$
& $72.54$
& $86.67$
& $5.28$ \\
\midrule

\multirow{4}{*}{128}
& MIME
& $\mathbf{36.14}$
& $\mathbf{73.33}$
& $\mathbf{85.79}$
& $\mathbf{6.35}$ \\
& TMR Early Fusion
& $34.04$
& $67.37$
& $78.51$
& $8.26$ \\
& TMR Late Fusion
& $30.26$
& $65.18$
& $77.63$
& $8.23$ \\
& LaMP
& $32.11$
& $66.32$
& $79.82$
& $7.87$ \\
\bottomrule
\end{tabular}%
}

\end{table}

MIME achieves the strongest performance across all hard-gallery sizes and evaluation metrics. Its advantage also generally increases as more semantically similar distractors are introduced. At a hard-gallery size of 128, MIME improves over TMR Early Fusion by 2.10 absolute points at R@1, 5.96 points at R@5, and 7.28 points at R@10, corresponding to relative improvements of 6.2\%, 8.8\%, and 9.3\%, respectively. MIME also reduces mean rank from 8.26 to 6.35, a relative reduction of 23.1\%. These results indicate that MIME more effectively separates motions whose captions share similar high-level semantics but differ in their underlying interaction details.

{
    \small
    \bibliographystylesupp{ieeenat_fullname}
    \bibliographysupp{sup}
}

\end{document}